\documentclass[conference]{IEEEtran}
\IEEEoverridecommandlockouts
\usepackage{cite}
\usepackage{amsmath}
\usepackage{amssymb}
\usepackage{amsfonts}
\usepackage{algorithmic}
\usepackage{graphicx}
\usepackage{textcomp}
\usepackage{xcolor}
\usepackage{multicol,multirow,lipsum}
\usepackage{mathtools}
\usepackage{cuted}
\DeclarePairedDelimiter\floor{\lfloor}{\rfloor}
\def\BibTeX{{\rm B\kern-.05em{\sc i\kern-.025em b}\kern-.08em
    T\kern-.1667em\lower.7ex\hbox{E}\kern-.125emX}}
\graphicspath{{./images/}}
\usepackage{listings}
\usepackage{xcolor}

\definecolor{codegreen}{rgb}{0,0.6,0}
\definecolor{codegray}{rgb}{0.5,0.5,0.5}
\definecolor{codepurple}{rgb}{0.58,0,0.82}
\definecolor{backcolour}{rgb}{1.0,1.0,1.0}

\lstdefinestyle{mystyle}{
    backgroundcolor=\color{backcolour},   
    commentstyle=\color{codegreen},
    keywordstyle=\color{blue},
    numberstyle=\tiny\color{codegray},
    stringstyle=\color{magenta},
    basicstyle=\ttfamily\footnotesize,
    breakatwhitespace=false,         
    breaklines=true,                 
    captionpos=b,                    
    keepspaces=true,                     
    numbersep=5pt,                  
    showspaces=false,                
    showstringspaces=false,
    showtabs=false,                  
    tabsize=2
}

\begin{document}

\title{A Deep Learning-based Multimodal Depth-Aware Dynamic Hand Gesture Recognition System}

\author{
    \IEEEauthorblockN{
        Hasan Mahmud, Mashrur M. Morshed, Md. Kamrul Hasan
    }
    \IEEEauthorblockA{
        Systems \& Software Lab (SSL), Department of Computer Science and Engineering (CSE) \\
        Islamic University of Technology (IUT) \\
        Board Bazar, Gazipur, Dhaka, Bangladesh \\
        \ttfamily\{hasan, mashrurmahmud, hasank\}@iut-dhaka.edu
    }
}

\maketitle

\begin{abstract}
The dynamic hand gesture recognition task has seen studies on various unimodal and multimodal methods. Previously, researchers have explored depth and 2D-skeleton-based multimodal fusion CRNNs (Convolutional Recurrent Neural Networks) but have had limitations in getting expected recognition results. In this paper, we revisit this approach to hand gesture recognition and suggest several improvements. We observe that raw depth images possess low contrast in the hand regions of interest (ROI). They do not highlight important fine details, such as finger orientation, overlap between the finger and palm, or overlap between multiple fingers. We thus propose quantizing the depth values into several discrete regions, to create a higher contrast between several key parts of the hand. In addition, we suggest several ways to tackle the high variance problem in existing multimodal fusion CRNN architectures. We evaluate our method on two benchmarks: the DHG-14/28 dataset and the SHREC’ 17 track dataset. Our approach shows a significant improvement in accuracy and parameter efficiency over previous similar multimodal methods, with a comparable result to the state-of-the-art.
\end{abstract}


\section{Introduction}
In our daily lives, we both consciously and subconsciously use numerous hand gestures. Human hands are dynamic and highly dexterous, allowing hands and hand movements to encode or represent a large variety of information. This capacity of hand gestures to represent information is second only to that of natural language. To account for physical disabilities related to speech, we use the sign language, where hand gestures play a significant role.


Formally, a hand gesture can be defined as the movement of the hands and fingers, in some particular orientation, with the intention of conveying meaningful information. This information can be something like some specific object (indicated by pointing fingers), or perhaps some intention (thumbs up indicating approval), or even specific symbols (fingers representing digits). Although many hand gestures are universal, they can also be culture or context specific. Symbolic gestures are generally static, that is, they exist only in the spatial domain. For example, the index finger representing the number $1$, or a singular quantity of something. This is a time invariant or static hand gesture (SHG). There are also dynamic hand gestures (DHG), which represent broader meanings---like moving the hand left and right to indicate refusal. Gestures like this work in both the spatial and temporal domains. We may think of such dynamic gestures as a sequence of static gestures, which together correspond to a single meaning.

Hand gesture-based interactions have long been introduced to many Human Computer Interaction (HCI) applications. Although their information embedding capacity is lower than that of natural language, speech controlled interaction has to consider the problem of vocal fatigue, or language barriers (for example, a person not knowing English may not be able to interact with their English-based system). In contrast, hand gestures can often be understood intuitively (such as pointing to a person or an object)---which is why people resort to gestures if there is a language barrier in communication. Hand gestures play a key role in many applications such as sign language communication, interacting with virtual objects in virtual environments, controlling robots, interacting with smart homes and ambient intelligence \cite{Survey_JI_2020, weise_2011, shotton_2011, HGRDesaiApplication_2017}. Thus, the development of robust hand gesture recognition systems can be considered as a key area of HCI research.


There are a wide variety of inputs to hand gesture recognition systems. Image or vision based inputs are the most ubiquitous and prevalent---including RGB images, stereo depth images, infrared based depth images, hand skeleton etc. Alternatively, there are less common approaches such as gloves embedded with accelerometers, gyroscopes, bend sensors, proximity sensors, and other forms of inertial sensors. The latter group of sensor approaches have limitations in terms of naturalness, cost, user comfort, portability, and data preprocessing---that is why vision based approaches are more widely used. 

Much like in other vision applications, deep learning has had a tremendous impact on the hand gesture recognition field. Initially, convolutional neural networks (CNN) were used, but they did not excel at recognizing dynamic gestures, which are spatio-temporal in nature. Later, researchers focused on recurrent neural network (RNN) based approaches---sometimes in conjunction with CNNs---which showed more success, as RNNs are capable of temporal modelling.

Even considering just vision approaches, there are several input modalities. This provides the unique opportunity to study multimodal input methods in the hand gesture recognition task. In this paper, we study the multimodal approach of using depth maps and 2D hand skeleton coordinates, both collected with infrared depth cameras. While both modalities provide spatio-temporal features, we obtain spatial depth information from depth-maps only.

\begin{figure}[htbp]
\centering
\includegraphics[width=0.45\linewidth]{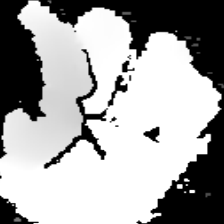}%
\includegraphics[width=0.45\linewidth]{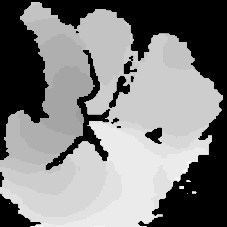}
\caption{(left) Original image (right) Image with quantized depth levels}
\label{fig_hand_1}
\end{figure}

We believe that rather than the exact depth value, CNN-based hand gesture recognition can learn more useful features from the relative depth between the different key areas of the hand, such as the finger, palm, or knuckle. As such, we propose a transform which does just this---quantizing continuous depth values to discrete depth 'levels' or 'regions', with a sharp contrast between adjacent levels. Figure \ref{fig_hand_1} shows a comparison between a depth and a quantized depth hand ROI.

Previously, Lai et al. \cite{kennethLai_2018} studied a CRNN approach to dynamic hand gesture recognition, by fusing features extracted from $227\times227$ depth maps and 2D hand skeleton coordinates. They obtained an accuracy of $85.46\%$ for 14 gestures, and $74.18\%$ for 28 gestures, on the DHG-14/28 benchmark \cite{dsmedt_CVPR_2016}. There hasn't been any further exploration of research in this direction. Instead, researchers have studied the usage of 3D hand skeleton coordinates \cite{NUNEZ201880}, graph based methods \cite{chen2019}, point-cloud approaches \cite{Min_2020} and so on, obtaining significantly better results. 

While theoretically sound, the multimodal fusion CRNN method of Lai et al. \cite{kennethLai_2018} suffers from high variance as the DHG-14/28 dataset is a relatively small dataset, while their proposed model is over-parameterized and lacks sufficient regularization.

It is not necessary to learn any fine surface texture features for dynamic hand gesture recognition. Thus, recognition performance should not be affected by using depth-map frames of lower resolution, which also has the added benefit of reducing the number of network parameters. We show in our work that in addition to quantizing the depth maps, we can also reduce the depth map resolution to as low as $50\times50$. We also add several regularization methods with the intention of increasing the robustness of our approach.


We can summarize our contribution points as follows:

\begin{enumerate}
    \item We provide a pre-processing method, Grayscale Variation ($gVar$), for hand depth-map frames, which quantizes depth values into discrete gray-levels with sharp intermittent contrast. The increased contrast between the finger and the palm regions improves gesture recognition performance.
    
    \item We provide a multimodal, feature-level fusion CRNN for dynamic hand gesture recognition, based on quantized depth values and 2D skeleton coordinates. Our approach shows several improvements over prior similar methods, and achieves 90.21\% and 88.64\% accuracy for 14 and 28 gestures respectively on the DHG-14 dataset \cite{dsmedt_CVPR_2016}, and 93.33\% and 
    90.24\% on the SHREC-2017 track dataset \cite{desmedt_2019}.
    
    \item Our method utilizes depth frames of much lower resolution compared to previously researched multimodal fusion CRNNs \cite{kennethLai_2018}, and has 8 times fewer parameters---while also having an increase in accuracy by  4.75\% and 14.45\% in DHG-14 and DHG-28 respectively.

\end{enumerate}

\section{Related Work}

Before the commercialization of depth-aware sensors, the predominant type of input to hand gesture recognition systems was color or RGB data, due to their relative ubiquity. Early approaches often used color-heuristics, such as Iwai et al. \cite{iwai1996gesture}, who utilized colored gloves along with decision trees to perform gesture recognition.

De Smedt et al. \cite{desmedt_2019} extracted effective descriptors from 3D skeleton coordinates by exploiting the geometric shape of the hand---an approach they termed as Shape of Connected Joints (SoCJ). The descriptors were encoded by a Fisher Vector representation, obtained via a Gaussian Mixture Model. A temporal pyramid was used to obtain the final features, which were fed into a linear SVM.

Nagi et al. \cite{nagi2011max} was one of the earliest researchers to suggest the usage of max-pooling based CNNs for hand gesture recognition. Later, authors in \cite{CNN_ASL_2018} used a typical CNN model for recognizing Kinect sensor-based American Sign Language (ASL).  

Conventional 2D CNNs have a limitation that they can only extract spatial information. To extract temporal information in dynamic gestures, three approaches are used, namely, 3D-CNN, two-stream networks, and Recurrent Neural Network (RNN)-based networks \cite{SHI_DHGR_2021}. In particular, Long-Short Term Memory (LSTM) networks---a type of RNN capable of learning long-term dependencies---have been found to be more suited for gesture recognition \cite{SHI_DHGR_2021}.

A dynamic gesture recognition system based on CNN and LSTM network is presented in \cite{CNN_LSTM_2017} based on Leap Motion Controller (LMC) device. Nunez et al. \cite{NUNEZ201880} proposed a unimodal 3D CNN + LSTM based approach with 3D skeleton coordinates, for generalized 3D human action recognition. They verified their method on several related sub-domains, including dynamic hand gesture recognition.
 
Researchers have also studied multimodal systems for gesture recognition. In \cite{MFFs_2018}, optical flow graph and color map information were combined to recognize dynamic hand gesture based on Jester \cite{jesterdataset_2019}, ChaLearn \cite{ChaLearnLA_2016}, and NVIDIA \cite{nvidia_2016_CVPR} datasets. Combination of RGB image, optical flow and depth information were used in \cite{Miao_2017_ICCV} to recognize dynamic gesture of human upper bodies.

Lai et al. in \cite{kennethLai_2018} proposed a multi-modal, 2D-CNN and LSTM based method, which we build upon in our work. Their approach utilized depth images and 2D-skeleton joint points for dynamic hand gesture recognition on the DHG-14/28 dataset \cite{dsmedt_CVPR_2016}.

Motion Feature Augmented Network (MFA-Nets) has been proposed in \cite{MFA-Net_2019}, where the authors exploit the finger motion features as well as global motion features from the hand skeleton data. These features are extracted using a variational autoencoder, and are passed along with regular features into an RNN. Köpüklü et. al. in \cite{MFFs_2018} follow a similar idea in their work Motion Fused Frames. They propose a data level fusion method, where motion information is incorporated into the static frames. In \cite{Hasanahci,airwriting}, the researchers fused depth features with depth quantized features for classification of hand air-writing activity.

In existing research, multimodal data fusion can thus be categorized into four stages \cite{multimodality_main_2021}: data-level\cite{MFFs_2018}, feature-level \cite{Miao_2017_ICCV,kennethLai_2018}, score-level \cite{kennethLai_2018}, and decision-level \cite{kennethLai_2018} fusion.

Apart from CNN and LSTMs, graph-based methods are also remarkably suitable for skeleton-based gesture recognition. The authors in \cite{Yan46} proposed a spatio-temporal graph convolutional network (ST-GCN). Chen et al. \cite{chen2019} proposed a novel dynamic graph-based spatio-temporal attention method for hand gesture recognition. Their method involves constructing a fully-connected graph from 3D skeleton joint coordinates, followed by learning the node features and edges by a self-attention mechanism.

Min et al. \cite{Min_2020} used point-clouds, generated from depth frames. They proposed the PointLSTM, a novel architecture designed to propagate information from past to future while preserving the spatial structure.

\section{Proposed Method}

Our proposed system consists of: (1) quantization of depth values into discrete gray levels, and (2) a strongly regularized, multi-modal CRNN architecture which takes in a sequence of depth image frames and a corresponding sequence of 2D skeleton joint points as input for the DHG task.


\begin{figure}[htbp]
\centering
\includegraphics[width=\linewidth]{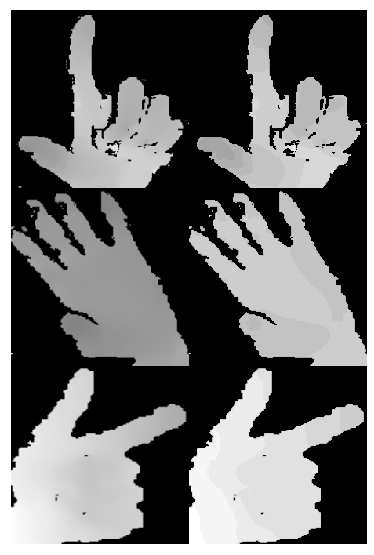}
\caption{(left) Original (right) Grayscale Variation}
\label{fig_gvar}
\end{figure}

\subsection{Grayscale Variation (gVar)}

The depth images in the SHREC-DHG-14/28 dataset \cite{3Dobject_2017} are 16-bit images. Of the available pixel range, only a very small portion is actually used by the hand gestures. Furthermore, the hand ROI does not possess enough contrast to highlight some features which may be useful to the recognition process, specially for fine gestures. For instance, information about finger position, orientation, overlap and motion (over multiple frames) can be useful in fine-grained gesture recognition. We thus address this issue in our preprocessing method, termed Grayscale Variation ($gVar$), which aims to increase the contrast in the hand region of interest.

The $gVar$ operation is a point-wise transform, and can be formulated as:

\begin{equation}
\begin{aligned}
    f(x, y) = G_{\min} + \floor*{
                    \left(
                        \frac{D(x, y) - D_{\min}}{D_{\max} - D_{\min}} \,\eta
                    \right) + 0.5}
                \\
                \times \floor*{
                    \frac{G_{\max} - G_{\min}}{\eta}
                }
    \label{eqn_1}
    \end{aligned}
\end{equation}

Where $f(x, y)$ denotes an output pixel, and $D(x, y)$ denotes an input pixel from the input depth image. We also utilize some hardware-specific information in $gVar$. The operation is only applied to non-zero pixels $> D_{near}$, where $D_{near}$ is the \textit{minimum optimal depth range} for the depth sensor used to collect the data. For the \textit{Intel RealSense SR300} used to collect the DHG-14/28 and the SHREC'17 datasets, $D_{th}$ is $200$ (0.66ft / 0.2m) \cite{sr300}.

From a high level overview, the Grayscale Variation operation reassigns depth values into $\eta$ discrete buckets, or \textit{gray levels}, in the closed range $[G_{\min}, G_{\max}]$. This creates several sharply contrasted regions in the depth-map, which highlights useful features such as the finger-tip and finger overlap. Moreover, the amount of output contrast is also controllable using the input parameters.

A very low value of $G_{min}$ would make it difficult to distinguish the hand from the background, and if $G_{min}$ and $G_{max}$ are not sufficiently spaced apart, the range of possible values would be compressed (and thus not have as much contrast as intended). The choice of $\eta$ also affects the quality of the output. For example, if we use too many gray levels, such as $\eta = 256$ (like a standard grayscale image), we would not obtain any useful contrast. However, if we use too few gray-levels, like perhaps 2-4, we may lose a significant amount of spatial and depth information. Empirically, a well-balanced choice of $G_{min}$ and $G_{max}$ are 155 and 255 respectively, with $\eta = 10$ levels between them.

We apply equation \eqref{eqn_1} on our input depth hand ROIs and get quantized grayscale hand ROIs, which are supplied to our model as inputs alongside 2D skeleton joint points.

\subsection{Proposed Architecture}

\begin{figure}[ht]
\centering
\includegraphics[width=0.8\linewidth]{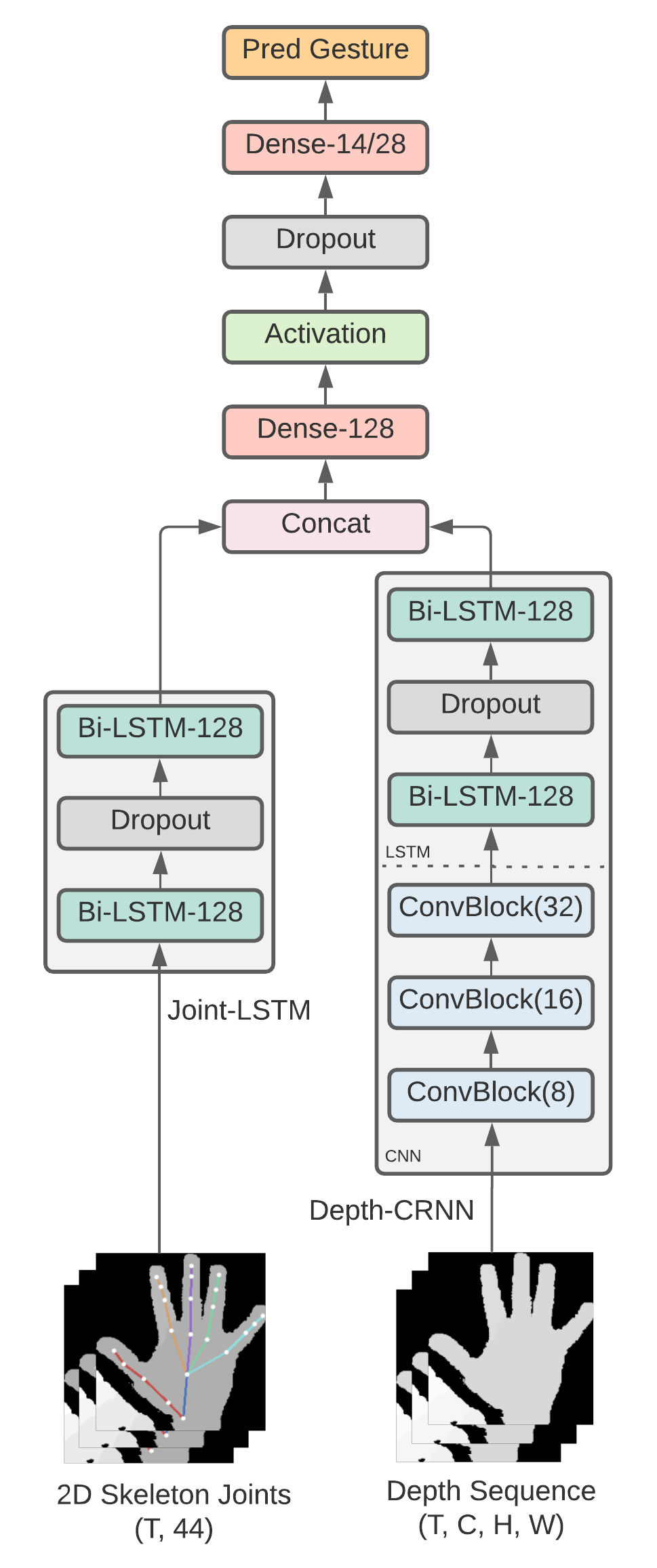}
\caption{Overall Fusion Model architecture, consisting of Joint-LSTM (left sub-network) and Depth CRNN (right sub-network). Features are fused by concatenation.}
\label{fig_fusion}
\end{figure}

Our base architecture follows the multi-modal fusion network proposed in \cite{kennethLai_2018}, and contains two main sub-networks: (1) A CNN + LSTM which processes depth image sequences, and (2) an LSTM network which processes 2D skeleton joint points. The feature maps produced by the two sub-networks are concatenated, before passing through an MLP head. Figure \ref{fig_fusion} shows the overall model architecture.

\subsubsection{Depth CRNN (CNN + LSTM)}
\begin{figure}[ht]
\centering
\includegraphics[width=0.55\linewidth]{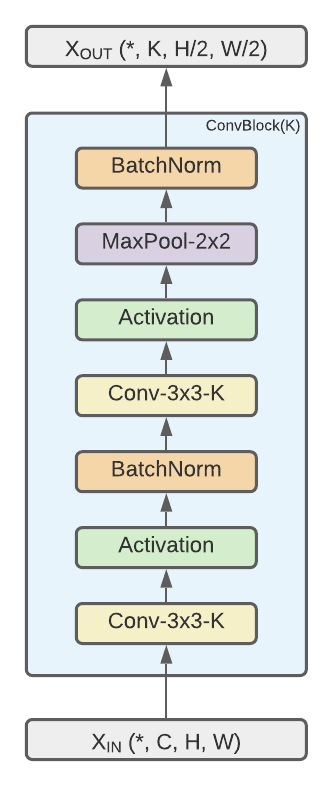}
\caption{Convolutional Block used to process depth sequence. Each $ConvBlock(K)$ produces output feature maps with $K$ channels.}
\label{fig_convblock}
\end{figure}

The CNN component contains three convolutional blocks, denoted as $ConvBlock$ in figure \ref{fig_fusion}. Each $ConvBlock$ is a sequence of Conv2D$_{3\times3}$, SiLU, BatchNorm, Conv2D$_{3\times3}$, SiLU, MaxPool, BatchNorm. This has also been visually shown in Figure \ref{fig_convblock}.

This is in fact, quite similar to the convolutional blocks used in \cite{kennethLai_2018} --- except that Lai et al. did not use any batch normalization layers, and we replace ReLU with SiLU / Swish activation \cite{swish}.

An arbitrary input depth image sequence may be a five-dimensional tensor of shape $(B, T, C, H, W)$ --- where $B$ represents batch size, $T$ represents the time-step size or sequence-length, and $C, H, W$ represent the channel, height and width resolutions of the images respectively. This five-dimensional tensor is rearranged into a four dimensional tensor of the form $(B \times T, C, H, W)$ and passed to the 2D convolutional layers for learning spatial features.

Since we are performing consecutive convolutions of a large batch size, $B * T$, we believe it is necessary to add batch normalization to the CNN component. Each BatchNorm layer is placed immediately prior to Conv2d layers, as their intended purpose is to normalize inputs to the convolutions. In our ablation studies (Section \ref{ablation}), we show that adding batch normalization has a very significant effect on model performance.

The feature-maps produced from the CNN component are reshaped to the form $(B, T, features)$ before being passed to the LSTM component. Previous fusion networks in \cite{kennethLai_2018} used a two layered LSTM with 256 hidden units. We instead propose using a two layered, bi-directional LSTM, each with 128 hidden units. This actually has the same output feature resolution as the regular LSTM with 256 units, but with the added benefit of lesser parameters and learning temporal features in both forward and backward directions.

Lastly, we add additional regularization, in the form of a dropout layer with a probability of 0.5, between the LSTM layers.

\subsubsection{Joint LSTM}
While the CNN is responsible for extracting spatial features, the LSTM component is focused on extracting temporal features. The joint based LSTM network is the same as the LSTM component of the other sub-network --- a two layered bi-directional LSTM with 128 hidden units and an intermediate dropout layer with a probability of 0.5.

\subsubsection{Feature Fusion}
The resultant features from the two sub-networks are fused (by concatenation), and passed to the MLP classifier. The overall process is shown in figure \ref{fig_fusion} as well as simplified in pseudocode \ref{pseudocodeffusion}. The researchers in \cite{kennethLai_2018} used MLPs with three fully connected layers, consisting of 256, 512 and 256 units, before the classification head with 14/28 units (depending on the no. of classes). We believe this had a significant contribution to the high variance problem in \cite{kennethLai_2018}.

\renewcommand{\lstlistingname}{Pseudocode}
\begin{lstlisting}[language=Python, caption={Einops \cite{einops} styled pseudocode for feature fusion model.}, label=pseudocodeffusion]
def feature_fusion(x1, x2):
    # x1: 3 dimensional input 2d-skeleton coordinate sequence
    # x2: 5 dimensional input depth-map sequence
    x1 = joint_lstm(x1)
    
    timesteps = x2.shape[1]
    x2 = rearrange(x2, "b t c h w -> (b t) c h w")
    x2 = depth_cnn(x2)
    x2 = rearrange(x2, "(b t) c h w -> b t (c h w)",
                  t=timesteps)
    x2 = depth_lstm(x2)
    
    x = concatenate([x1, x2], axis=1)
    x = rearrange(x, "b t f -> b (t f)")
    x = mlp(x)
    
    return x
\end{lstlisting}

In our model, we reduce the MLP to having just a single fully connected layer with 128 units before the classification head. We also added a dropout layer with probability 0.5 after the fully connected layer, to further regularize the model. A comparison of the number of total model parameters are shown in table \ref{table_params}.

\begin{table}[h]
	\centering
	\caption{Trainable Parameters}
	\label{table_params}
	\resizebox{\linewidth}{!}{
	\begin{tabular}{|c|c|c|}
		\hline
		Method & Depth Resolution & \#Params \\ 
		\hline
		FL-Fusion-Concat \cite{kennethLai_2018} & $227\times227$  & $31.7$ M \\
		\hline
		SL-Fusion-Avg \cite{kennethLai_2018}    & $227\times227$  & $32$ M \\
		\hline
        gVar-FL-Fusion (ours)   & $50\times50$    & $4.4$ M \\
		\hline
	\end{tabular}
	}
\end{table}

\section{Datasets}
\label{datasets}

We evaluate our method on two datasets, the SHREC-2017 track dataset \cite{desmedt_2019} and the DHG-14/28 dataset \cite{dsmedt_CVPR_2016}.

\begin{table}[ht]
    \centering
    \caption{Gesture Recognition Classes in the DHG-14/28 Dataset}
    \label{tbl_classes}
    \resizebox{0.8\linewidth}{!}{
    \begin{tabular}{|c|c|c|c|}
        \hline
        Class & Gesture & Short & Grain \\
        \hline
        0 & Grab  & G & Fine \\
        1 & Tap  &  T & Coarse \\
        2 & Expand & E & Fine \\
        3 & Pinch & P & Fine \\
        4 & Rotation Clockwise  & R-CW & Fine \\
        5 & Rotation Counter-clock & R-CCW & Fine \\
        6 & Swipe Right  &  SR & Coarse \\
        7 & Swipe Left  &  SL & Coarse \\
        8 & Swipe Up  &  SU & Coarse \\
        9 & Swipe Down & SD &  Coarse \\
        10 & Swipe X  & SX & Coarse \\
        11 & Swipe V  & SV &  Coarse \\
        12 & Swipe +  & S+ & Coarse \\
        13 & Shake  & Sh & Coarse \\
        \hline
    \end{tabular}
    }
\end{table}

\begin{figure}[htpb]
\centering
\includegraphics[width=8cm]{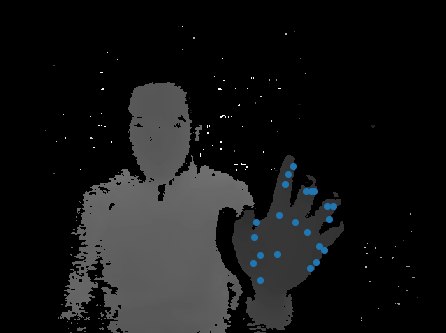}
\caption{Depth Image frame with Corresponding Joint Point}
\label{fig_datatrue}
\end{figure}

\subsection{DHG-14/28 Dataset}
The DHG-14/28 dataset consists of 14 types of dynamic hand gestures, performed two ways: with one finger and with two fingers. The gestures are performed by 20 different people, where each person repeats a gesture 5 times. This leads to a total of 2800 gesture instances. There are three modalities in the dataset. Each hand gesture instance consists of a sequence of depth image frames, a sequence of 2D hand-skeleton joint points, and a sequence of 3D hand-skeleton joint points. For each sequence, the effective start and end of the gesture is also marked. There are primarily 14 target gesture labels. When one finger and two finger gestures are considered to be separate classes, there are 28 gesture classes. These gestures are further categorized into fine and coarse grained gestures, which can be seen from table \ref{tbl_classes}. 

\begin{table*}[ht]
	\centering
	\caption{Detailed Accuracy Comparison between Fusion Methods for DHG-14}
	\label{tbl_fusion_acc}
	\resizebox{\textwidth}{!}{%
	\begin{tabular}	{ |c|c|c|c|c|c|c|c|c|c| }
		\hline
		\multirow{2}{*}{Method} & \multicolumn{3}{c|}{Fine} & \multicolumn{3}{c|}{Coarse} & \multicolumn{3}{c|}{Both} \\
		\cline{2-10}
		
		& Best & Worst & Avg $\pm$ Std
		& Best & Worst & Avg $\pm$ Std
		& Best & Worst & Avg $\pm$ Std \\
		\hline
		FL-Fusion-Concat \cite{kennethLai_2018} 
		& 90 & 48 & 72.90 $\pm$ 10.30
		& 98.89 & 78.89 & 86.83 $\pm$ 4.68
		& 87.86 & 67.86 & 81.86 $\pm$ 5.38 \\
		\hline
		SL-Fusion-Avg \cite{kennethLai_2018} 
		& 92 & 52 & 76 $\pm$ 10.51
		& 97.78 & 81.11 & 90.72 $\pm$ 4.64
		& 95 & 72.86 & 85.46 $\pm$ 5.16 \\
		\hline
		gVar-FL-Fusion (ours)
		& \textbf{96} & \textbf{64} & \textbf{82.6 $\pm$ 8.28}
		& \textbf{100} & \textbf{82.22} & \textbf{94.44 $\pm$ 4.58}
		& \textbf{96.43} & \textbf{80.00} & \textbf{90.21 $\pm$ 5.06} \\
		\hline
	\end{tabular}
	}
\end{table*}

\subsection{SHREC 2017 Track Dataset}
The SHREC 2017 Track dataset is similar to the DHG-14/28 dataset in some ways. It consists of the same gesture classes (14 and 28) and the method of collecting the data is similar. There is no effective start and end for the SHREC dataset --- we can consider the entire gesture sequence as relevant. A total of 2800 data instances are available, divided into a 70\% training set and a 30\% test set. Furthermore, the training set contains data from 20 performers, while the test set contains data from an additional 8 unseen performers.

\section{Experiment Details}
\label{experimental-details}

\subsection{Data Preprocessing}
First, we extracted all the hand region of interests (ROI) from the depth image frames, from both datasets. For the DHG-14/28 dataset, only the depth frames between the effective start and effective end of the gesture are used.

We apply the $gVar$ transform described in equation \ref{eqn_1} and resize all the images to $50\times50$ resolution. For the joint points, we normalize them by subtracting the position of the palm point of the first joint frame from the rest. Similar to the depth image frames, we only take the joint points between the effective start and end of the gesture for the DHG-14/28 dataset.

\subsection{Augmentation}
Since the number of instances in the DHG-14/28 dataset and the SHREC-2017 track dataset are quite low, it is important to apply some augmentation techniques to prevent overfitting. We apply the following techniques:

\subsubsection{Time Shift}
We apply a time shift of 0.1, with a probability of 0.5. That is, the image and joint sequences are shifted randomly within a range of $\pm10\% $ (of the total number of frames). As we use sequences with length 32, the gestures would be shifted left or right by at most 3 frames.

\subsubsection{Shift-Scale-Rotate}
We apply random shift, random scale and random rotate to the depth image and joint points, with a range of $\pm0.2$, $\pm0.2$, and $\pm20^{\circ}$ respectively. Furthermore, these transforms are also applied with a probability of 0.5. It is to be mentioned that the gestures $R-CW$ and $R-CCW$ (rotate clockwise and counterclockwise) are rotation sensitive. However, we empirically found that a rotation of only $\pm20^{\circ}$ simply acts as jitter to the gesture, and has some minute contributions to generalization.

The shift-scale-rotate transforms are applied uniformly throughout the gesture sequence --- i.e. all frames are shifted by $p$, scaled by $q$ and rotated by $r^{\circ}$. While it is trivial to apply these transforms to images, it is a bit trickier to apply them to coordinates. We compute the rotation matrix for each joint point frame by considering that frame's palm point coordinate as the center.

\subsection{Training Details}
We follow the standard experimental method described in \cite{dsmedt_CVPR_2016, kennethLai_2018, kenneth_ensemble_2020, chen2019, NUNEZ201880} for the DHG-14/28 dataset. We perform a twenty-fold cross validation. Among the 20 subjects present in the dataset, in each iteration, 19 subjects are used for training while the model is evaluated on the remaining 20th subject. This is a highly robust method of evaluating hand gesture recognition methods, as a model always needs to generalize to an unseen distribution in each fold.

Our model takes corresponding pairs of depth image frames and 2D skeleton joint points as input. We use a batch size of 16 and a time-step size of 32. Exactly 32 frames are used from a given data instance --- however, it is possible for data items to contain both more and less than 32 frames. As such, for sequences with less than 32 frames, we pad with blank frames, and for sequences with more than 32 frames, we perform evenly distributed sampling between the start and end frame. Each image frame is resized to 50x50 resolution and grayscale variation ($gVar$) is applied.

We use the AdamW optimizer \cite{adamw}, with an initial learning rate of 0.0003. Furthermore, we apply warm-up scheduling for the first 10 epochs, followed by cosine annealing scheduling for the remaining epochs. For additional regularization, we also use weight decay of 0.1 and label smoothing of 0.1.

Identical hyperparameters and settings are used for the SHREC 2017 track, except without 20-fold cross validation. The model is trained and tested on the provided 70-30 split, as done in \cite{desmedt_2019, chen2019, Min_2020}.

\section{Experimental Results}
\label{experimental-results}

\subsection{Comparison With Previous Fusion Methods}
At its core, our work proposes several improvements over the fusion architecture proposed by Lai et al. \cite{kennethLai_2018}. As such, we first go into a detailed comparison between the previously researched fusion architecture and ours.

From Table \ref{tbl_fusion_acc} we can see that gVar-FL-Fusion shows an undisputed improvement over previous fusion methods. There is more improvement in the recognition of fine gestures (+6.6\%), compared to coarse gestures (+3.72\%).

\subsection{Comparison With Existing Methods on DHG-14/28}
Our method shows comparable performance to the start-of-the-art on the DHG-14/28 dataset. Since the DHG dataset is evaluated with a difficult 20-fold cross validation process, results obtained on DHG are quite robust.

From Table \ref{tbl_acc_1}, we can see that while 14 gesture recognition improved by almost $5\%$, there is an astounding improvement of more than $14\%$ in the recognition rate of 28 gestures, in comparison to previously researched fusion models (SL-Fusion-Avg \cite{kennethLai_2018}).

\subsection{Comparison With Existing Methods on SHREC-2017}

Although the depth and 2D joint fusion models in \cite{kennethLai_2018} were not previously evaluated on the SHREC-2017 dataset, for completeness, we also ran experiments on this benchmark.

From Table \ref{tbl_acc_2} we can see that while our method is not comparable to the state-of-the-art, it still shows competitive results. However this also alludes to the fact that future research should consider point clouds as an input modality, as the performance of point cloud based models notably outperform all other existing methods. 

\section{Ablation}
\label{ablation}

Although it is quite difficult to perform a complete grid comparison to find out the exact usefulness of each design choice, we can obtain an approximate idea with leave-one-setting-out experiments. This means, from some baseline settings, we will change only one specific setting in an arbitrary run. We run this experiment over the SHREC'17 track dataset (14 gestures), in the following ways:

\begin{enumerate}
    \item Reported settings in Section \ref{experimental-details}
    \item Without using $gVar$
    \item Without using dropout
    \item Without using BatchNorm
    \item Using ReLU instead of Swish/SiLU
    \item Doing all of the above
\end{enumerate}

Table \ref{tbl_ablation} shows the results of our experiments. From the results, we can infer that all these choices have a definite impact on the model's performance, but in different degrees. It seems as though the absence of BatchNorm layers most critically affect the model, while the usage of SiLU instead of ReLU provides a minor boost of 0.4\% accuracy. $gVar$ also has a very significant effect---it results in a 1.43\% improvement.

\begin{table}[ht]
	\centering
	\caption{Evaluation of the effect of different factors (on SHREC 14 gestures)}
	\label{tbl_ablation}
	\resizebox{0.95\linewidth}{!}{%
	\begin{tabular}	{ |c|c|c|c|c|c| }
		\hline
		Activation & BatchNorm & Dropout & $gVar$  & Accuracy \\ 
		\hline
		SiLU & $\checkmark$ & $\checkmark$ & $\checkmark$ & \textbf{93.33} \\
		\hline
		SiLU & $\checkmark$ & $\checkmark$  & $\times$ & 91.9 \\
		\hline
		SiLU & $\checkmark$ & $\times$  & $\checkmark$ & 92.38 \\
		\hline
	    SiLU & $\times$ & $\checkmark$   & $\checkmark$ & 90.71 \\
		\hline
		ReLU & $\checkmark$ & $\checkmark$  & $\checkmark$ & 92.98 \\
		\hline
		ReLU & $\times$ & $\times$  & $\times$ & 89.64 \\
		\hline
	\end{tabular}
	}
\end{table}

\begin{table*}[ht]
	\centering
	\caption{Accuracy Comparison on the DHG-14/28 Dataset}
	\label{tbl_acc_1}
	\resizebox{0.8\linewidth}{!}{%
	\begin{tabular}	{ |c|c|c|c|c| }
		\hline
		Method & Fine-Grain & Coarse-Grain & 14-Gestures & 28-Gestures \\ 
		\hline
		FL-Fusion-Concat \cite{kennethLai_2018} & 72.9   & 86.83   & 81.86   & - \\
		\hline
		SL-Fusion-Avg \cite{kennethLai_2018}    & 76     & 90.72   & 85.46   & 74.19 \\
		\hline
        3D\_CNN + LSTM \cite{NUNEZ201880}   & 78     & 89.8    & 85.6    & 81.1 \\
        \hline
        MFA-Net \cite{MFA-Net_2019}          & 75.6   & 91.3    & 85.75   & 81.04 \\
        \hline 
        Ensemble of Models \cite{kenneth_ensemble_2020}         & 81.2   & 88.8    & 86.11   & - \\
        \hline
        DG-STA \cite{chen2019}           &  -    &  -       & \textbf{91.9}    & 88 \\
        \hline
        gVar-FL-Fusion (ours)   & \textbf{82.6}     & \textbf{94.94}       & 90.21   & \textbf{88.64} \\
		\hline
	\end{tabular}
	}
\end{table*}

\begin{table*}[ht]
	\centering
	\caption{Accuracy Comparison on the SHREC-2017 Dataset}
	\label{tbl_acc_2}
	\resizebox{0.8\linewidth}{!}{%
	\begin{tabular}	{ |c|c|c|c| }
		\hline
		Method & Modality & 14-Gestures & 28-Gestures \\ 
		\hline
		Key Frames \cite{3Dobject_2017} & Depth Sequence & 82.9 & 71.9 \\
        \hline
        SoCJ+HoHD+HoWR \cite{dsmedt_CVPR_2016} & 3D-Skeleton & 88.2 & 81.9 \\
        \hline
        Res-TCN \cite{res-tcn} & 3D-Skeleton & 91.1 & 87.3 \\
        \hline
        MFA-Net \cite{MFA-Net_2019} & 3D-Skeleton & 91.31   & 86.55 \\
        \hline 
        ST-GCN \cite{Yan46} & 3D-Skeleton & 92.7 & 87.7 \\
        \hline 
        DG-STA \cite{chen2019} & 3D-Skeleton &  94.4    &  90.7 \\
        \hline
        PointLSTM \cite{Min_2020} & Point Clouds & \textbf{95.9} & \textbf{94.7} \\
        \hline
        gVar-FL-Fusion (ours)   & Depth Sequence + 2D-Skeleton & 93.33    & 90.24\\
		\hline
	\end{tabular}
	}
\end{table*}
\section{Result Analysis}
\label{result-analysis}

It is apparent from Table \ref{table_params} and \ref{tbl_fusion_acc} that despite using significantly fewer parameters and inputs at much lower resolutions, our proposed system shows significant improvement in generalization on the DHG-14 dataset. The average performance of our feature level fusion setup shows an improvement of $\sim$8\% over the feature level fusion, and $\sim$5\% over score level fusion, shown in \cite{kennethLai_2018}.

More importantly, it was previously noted in \cite{kennethLai_2018} that multimodal fusion models struggle to perform on 28 gestures, as they obtained $85.46\%$ and $74.19\%$ on 14 and 28 gestures respectively (also observable in Table \ref{tbl_acc_1}. The authors in \cite{kennethLai_2018} mentioned that the fusion networks were not optimized for 28 gesture recognition. From our results, it is apparent that we have overcome this limitation of fusion models. We argue that the reasoning behind this improvement is a combination of factors --- such as, a more strictly regularized and comparatively smaller model (thus being less prone to overfitting), the addition of normalization layers which address the previously ignored possible internal covariate shift, and useful preprocessing ($gVar$) which highlights the finger regions in depth images.

\begin{figure}[ht]
\centering
\includegraphics[width=\linewidth]{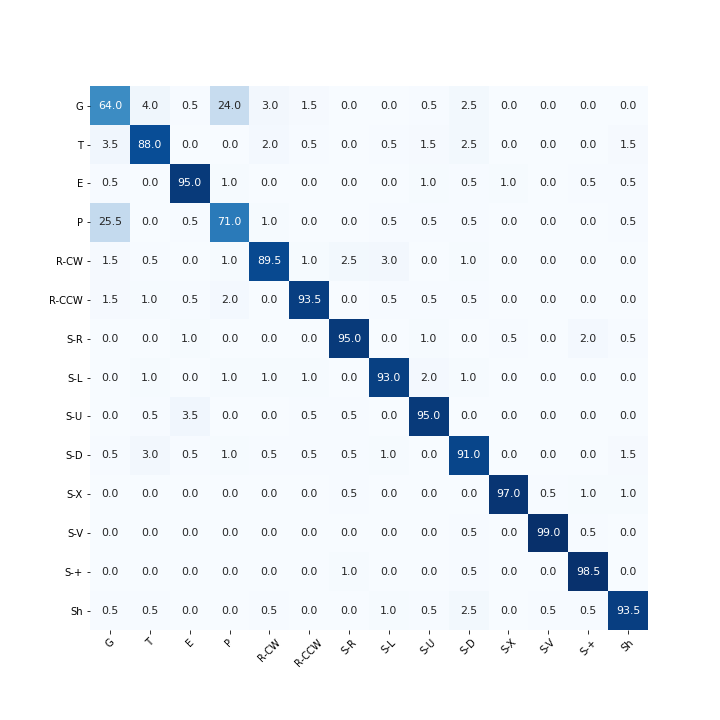}
\caption{Confusion matrix for proposed gVar-FL-Fusion}
\label{fig_confmat}
\end{figure}

In particular, it can be observed from figures \ref{fig_gvar} and \ref{fig_hand_1} and  that the $gVar$ operation reduces the contrast present in the hand ROI, and also highlights the fingers and extremities of the hand. We believe this representation of the depth images allows the model to extract some additional useful information, which is critical to the performance of fine gestures, where finger movements dominate the gesture sequence. Figure \ref{fig_confmat} shows the confusion matrix for the DHG-14/28 dataset. Although from table \ref{tbl_acc_1} we do see that our method has significantly improved results, the confusion matrix actually provides us with deeper insights. We can see that there is still some room for improvement --- the model particularly has some confusion between the Grab and the Pinch gesture. This phenomenon can also be observed in prior works, such as \cite{kennethLai_2018, kenneth_ensemble_2020, MFA-Net_2019}, since the difference between the Grab and the Pinch gesture is quite minimal.

\section{Conclusion}
\label{conclusion}

Our work is primarily focused on improving the multi-modal fusion approach to dynamic hand gesture recognition. From our experiments, it can be inferred that the primary reason for the significant performance gain is that we focused on improving the input representation, using rationale as well as hardware-specific information. Furthermore, previously researched fusion models were also highly over-parameterized, and several factors in the model design were not well thought out or rigorously investigated.

We hope that our work fuels further research in the field of multimodal dynamic hand gesture recognition.




\bibliographystyle{IEEEtran} 

\bibliography{Bibliography} 


\end{document}